\documentclass[10pt]{article} 
\usepackage[accepted]{rlc}

\usepackage{amssymb}            
\usepackage{mathtools}          
\usepackage{mathrsfs}           
\mathtoolsset{showonlyrefs}     
\usepackage{graphicx}           
\usepackage{subcaption}         
\usepackage[space]{grffile}     
\usepackage{url}                

\usepackage[noend]{algpseudocode}
\usepackage{textcomp}
\usepackage{xcolor}
\usepackage{svg}
\usepackage{subcaption}
\usepackage{multirow}
\usepackage{amssymb}
\usepackage{amsmath}
\usepackage{float}

\newtheorem{definition}{Definition}

\usepackage{algorithm}
\usepackage{algorithmicx}
\usepackage[noend]{algpseudocode}

\title{ ODGR: Online Dynamic Goal Recognition}

\author{Matan Shamir\thanks{These authors contributed equally.} \\
    matan.shamir@live.biu.ac.il \\
    Department of Computer Science \\
    Bar-Ilan University, Ramat Gan, Israel
    \And
    Osher Elhadad\footnotemark[1] \\
    osher.elhadad@live.biu.ac.il \\
    Department of Computer Science \\
    Bar-Ilan University, Ramat Gan, Israel
    \And
    Matthew E. Taylor  \\
    matthew.e.taylor@ualberta.ca \\
    Department of Computing Science \\
    Alberta Machine Intelligence Institute (Amii)\\
    University of Alberta, Edmonton, Canada
    \And
    Reuth Mirsky  \\
    mirskyr@cs.biu.ac.il \\
    Department of Computer Science \\
    Bar-Ilan University, Ramat Gan, Israel}

\begin{document}

\maketitle

\begin{abstract}
Traditionally, Reinforcement Learning (RL) problems are aimed at \textit{optimization} of the behavior of an agent. This paper proposes a novel take on RL, which is used to learn the policy of another agent, to allow real-time \textit{recognition} of that agent's goals.
Goal Recognition (GR) has traditionally been framed as a planning problem where one must recognize an agent's objectives based on its observed actions. Recent approaches have shown how reinforcement learning can be used as part of the GR pipeline, but are limited to recognizing predefined goals and lack scalability in domains with a large goal space. This paper formulates a novel problem, ``Online Dynamic Goal Recognition'' (ODGR), as a first step to address these limitations. Contributions include introducing the concept of dynamic goals into the standard GR problem definition, revisiting common approaches by reformulating them using ODGR, and demonstrating the feasibility of solving ODGR in a navigation domain using transfer learning. 
These novel formulations open the door for future extensions of existing transfer learning-based GR methods, which will be robust to changing and expansive real-time environments.
\end{abstract}

\section{Introduction}
 Goal Recognition plays a crucial role in a variety of domains in Human-Robot Interaction~\citep{massardi2020parc, jiang2021goal, shvo2022proactive, geib2002problem, inam2018risk} and Multi-Agent Systems~\citep{avrahami2005fast, freedman2017integration, su2023fast}.
 Reasoning about the goal of another agent presents an important challenge that complements an ego agent's policy learning and can inform and leverage RL techniques by facilitating significant breakthroughs in how machines perceive and interpret other agents, especially human intentions.
 
 Existing GR approaches commonly assume that a fixed set of goals is provided as part of the problem formulation~\citep{mirsky2021introduction, meneguzzi2021survey}. 
%
 In recent GR approaches that leverage Reinforcement Learning (RL), which are also called \textit{GR as RL}~\citep{amado2022goal},
 a policy is learned for each goal during a dedicated learning phase, often occurring offline before the GR task. Then, during the inference phase, the algorithm is given a sequence of observed actions and uses inference methods to perform GR. Unlike planning-based GR~\citep{ramirez2010probabilistic, meneguzzi2021survey}, these approaches have two fundamental limitations: first, all possible goals must be specified before the inference phase. This requirement may not hold in dynamic environments where goals can emerge or change over time. Second, an issue of scalability arises when dealing with domains that encompass many goals. GR requires significant computational effort and resources for considering each individual goal, posing a challenge for large-scale applications. This problem is further exacerbated in learning-based GR methods, where training an agent for each goal becomes prohibitively costly and infeasible due to the computational burden and time required for training.

Considering these two limitations, the first contribution of this paper is conceptual: it formulates a problem called \textit{Online Dynamic Goal Recognition} (ODGR) with a temporal setting, where the time steps in which inputs are given are specified.
By generalizing the definition of GR, this paper can identify the strengths and limitations of various GR frameworks across different settings and assumptions.
%
For example, consider the potential locations that a researcher attending the RLC 2024 conference will choose to visit during their day off. First, a recognition task should focus on navigation in Massachusetts's towns. Once inside Amherst, one might recognize the locations people are moving towards to identify which attraction they are interested in. Lastly, if the researcher is visiting the Beneski Museum of Natural History, the recognition task changes to recognize the specific exhibition that the researcher is interested in. One alternative is representing each task as a different recognition problem, yet this solution means that a new set of goals needs to be learned for each new task. This paper suggests a formulation where these tasks can all be captured using a single recognition problem with a single set of states and actions and changing goals.

The second contribution of this paper is algorithmic: we offer a general algorithm that leverages Transfer Learning between RL tasks and demonstrates the feasibility of a future framework that would implement this solution by presenting a proof-of-concept for simple navigational environments. The suggested algorithm expands the capacity of traditional learning-based GR methods by adapting to ``dynamic goals'' -- goals not set before the inference phase. 
For this simple navigational domain, we propose a heuristic method that aggregates existing Q-functions into new Q-functions for the dynamic goals, which can adequately facilitate recognition. 
In the proposed solution, existing Q-functions are learned before any observations are given, and they are specifically designed to represent the expected policies for a predefined set of base goals. The success of our solution is measured by the ability of the recognizer to use the generated policies when performing the recognition process and succeed in inferring the correct goal.

By introducing the concept of dynamic goals and presenting an approach for policy transfer in navigational domains, we show that GR methods can be extended to recognize and adapt to goals that emerge or change over time, thereby broadening the range of problems and real-time domains to which these methods can be applied. This research provides a significant first step towards a more robust and adaptable GR system capable of handling dynamic environments.

\section{Preliminaries}


The basic model used to describe a domain is a \textbf{Markov Decision Process (MDP)} M, which is a 4-tuple $\langle S, A, P, R \rangle$ such that $S$ represents a set of states in the environment, $A$ is the set of actions the agent can execute, $P$ is a transition function, and $R$ is a \textbf{reward} function. A \textbf{transition} function $P(s' | s, a)$ returns the probability of transitioning from state $s$ to state $s'$ after taking action $a$, and a reward function $R(s, a, s')$ returns the reward obtained when an agent transitions from $s$ to $s'$ using action $a$. 
The \textbf{utility function $Q$} (or Q-function) is a function $Q: S \times A \rightarrow \mathcal{R}$ that defines the expected utility of executing an action $a \in A$ in state $s \in S$. A Q-learning agent estimates the values of $Q(s, a)$ from its experiences. A \textbf{Policy} $\pi: S \times A \rightarrow  [ 0,1 ] $ represents the probability that an agent would choose to perform an action $a \in A$ in state $s \in S$. 




The Goal Recognition (GR) problem consists of two agents: an \textit{actor} and an \textit{observer}. The GR problem is defined from the observer's perspective, who needs to recognize the actor's goal. Recognizing someone's goal can be challenging, particularly when these goals undergo sudden and unanticipated transformations. 
This work relies on a line of work in which a GR algorithm produces a set of policies or behaviors, one per goal \citep{polyvyanyy2020goal, Ko2023plan, chiari2023goal}. We follow the utility-based GR problem formulation from \citet{amado2022goal}: 


\begin{definition}
A Goal Recognition problem is a tuple $\langle S, A, O, G, Q_G\rangle$, such that $S$ is a set of states and $A$ is a set of actions an observed actor can take, $O$ is a sequence of observations which are states and actions tuples, $G \subseteq S$ is a set of goals (the actor pursues only one goal at a time), and $Q_G$ is a set of Q-functions $\{Q_g\}_{g \in G}$. The output of a GR problem is a goal $g \in G$ that best explains $O$.
\end{definition}

\noindent Notice that this definition is very broad to enable different solutions. First, it does not force a specific way to learn the Q-functions that the actor will assume to reach $g$. Second, it does not elaborate on how a goal explains the observations. Third, there is no single metric to quantify the quality of one explanation over another.

Addressing GR through existing RL methodologies, such as those proposed by~\cite{amado2022goal}, involves learning a distinct policy for each goal within a predefined goal set and then comparing these policies against observed behaviors. Importantly, policies should not be perfectly optimal to tackle many GR challenges. 
Hence, this article will introduce the application of zero-shot transfer learning and heuristic strategies in simple navigational domains as a first step in developing sub-optimal policies that nevertheless provide precise GR outcomes.
This paper focuses on goal-directed reward functions (i.e., a reward function-oriented towards a goal $g$).  In addition, we focus on a specific implementation of GR using Q-Learning, but we highlight that the ODGR problem, in general, is not limited to this implementation. To represent the behavior of agents pursuing each potential goal in $G$, we use a set $\Pi_G$. 
Each policy $\pi_g$ for an actor pursuing goal $g$ is a softmax based on the Q values over all actions in state $s$: 
%
$\pi_g(a|s) = \frac{Q_g(s,a)}{\sum_{a' \in \mathcal{A}} Q_g(s, a')}$.
%
Then, the policy for an agent pursuing a goal $\pi_g$ chooses an action based on this probability distribution. 
Using this formulation, we can say that a goal $g$ explains an observation sequence by evaluating the likelihood that the actor will produce $o$ from $\pi_g$. 
As this paper focuses on learning dynamic goals, we use an existing underlying approach to learn $\Pi_G$ and follow a similar learning procedure as \citet{amado2022goal} to learn $\Pi_G$. 


\section{Online Dynamic Goal Recognition}
Certain methodologies for GR rely on the notion that inputs to the problem arrive at distinct time steps, along with varying assumptions regarding their nature, such as whether they constitute goals or new observations. For example, in Plan Recognition as Planning \citep{ramirez2010probabilistic}, it is expected that the description of the environment will be decoupled from a specific problem instance, and different GR problems can be defined for the same environment. We now present a formulation that aims to identify the different components that can change within each problem, potentially facilitating the selection of a more appropriate framework based on the expected scenarios the user may encounter, and the requirements from the expected solution.

\begin{definition} \label{ODGR}
An Online Dynamic Goals Recognition (ODGR) problem is a tuple 
$\langle T, \langle G^i, \{O\}^i \rangle_{i\in 1..n}\rangle$, 
where $T = \langle S, A \rangle$ is a domain theory such that $S$ is the set of all possible states and $A$ is the set of actions, $G^i$ is a set of goals such that $\forall i \in \{1..n\}, G^i\subseteq S$, and $\{O\}^i$ represents a set of observations sequences (which might contain gaps), where each $O_j \in \{O\}^i$ is a single observation sequence $O_j = \langle o_j^1, o_j^2 \ldots \rangle = \langle \langle s_j^1, a_j^1 \rangle, \langle s_j^2, a_j^2 \rangle, \ldots \rangle$. 
Aligning with other online problems, each input is given at an increasing time-step, lexicographically ordered, while it is inherent that because of the input dependency, $T, G^i$ and $O_j \in \{O\}^i$ arrive at increasing time-steps $\forall i, j$. For each two sequences of observations $o^l_t, o^m_t \in O_t$ $o^z_j, o^k_j \in O_j$ where $t < j$, $l < m$, and $z < k$, it holds that the observations are given chronologically, such that $o^l_t < o^m_t < o^z_j < o^k_j$.
\end{definition}

An algorithm for the ODGR problem is expected to return a goal $g\in G^i$ that best explains $O_j\in\{O\}^i$ upon its arrival for all $i,j$. The final output is a set of goals $G^* = \{\{g_1^1, g_2^1, ...\}, \{g_1^2, g_2^2, ...\}, ..., \{g_1^n, g_2^n, ...\}\}$ where $g_j^i$ is the goal returned by the algorithm upon receiving $O_j\in\{O\}^i$.
%
%
Using this formulation, one can now define a general GR \textit{domain} according to the domain theory $T$. In this work, the domain consists of an MDP, but the formulation allows for different representations as well. For each domain, several \textit{instances} can be constructed at different times according to the set of potential goals $G^i$ the actor may be pursuing. For each instance, a single recognition \textit{problem} is an observation trace that needs to be explained by $G^i$. We can split an ODGR problem into several time intervals, depending on the reception of each input type, $T$, $G$, and $O$. 

\begin{figure*}
  \centering
  \includegraphics[width=0.8\textwidth]{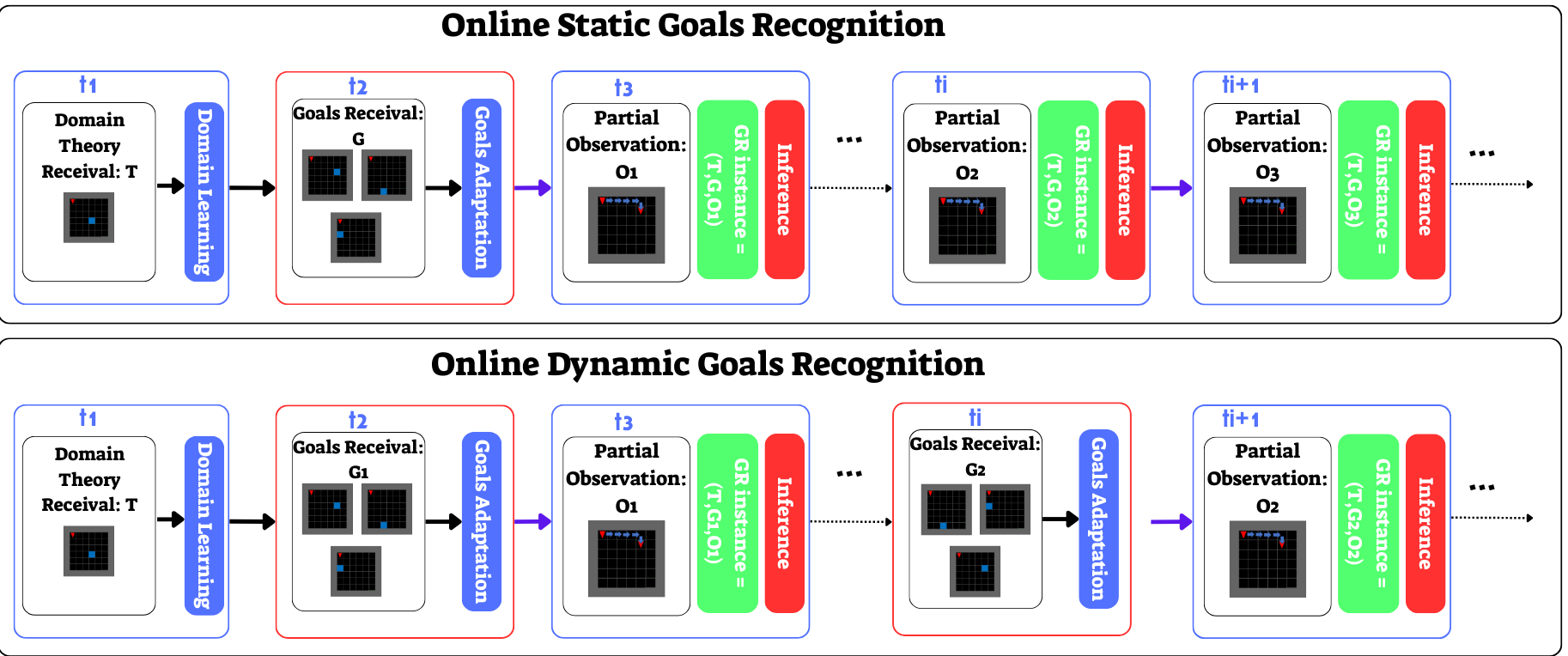}
  \caption{OSGR and ODGR. The symbol $t_i$ denotes the initiation time of each process.} 
  \label{fig:GRAML}
\end{figure*}
    \textbf{Domain Learning Time} is the duration necessary from receiving the domain theory $T$ until concluding the domain-specific processing and being prepared to receive $G$.
    
    \noindent \textbf{Goals Adaptation Time} is the duration from receiving $G$ until completing the inner-state changes and becoming ready to perform inference based on observation trajectories $O$.
    
    \noindent \textbf{Inference Time} is the time it takes to process an observation and produce a goal in response.


A GR framework can receive a new piece of input before completing the processing of the previous one. Following Allen's interval algebra \citep{allen1983maintaining}, we say that one input may \textit{overlap} another input. These definitions remain independent of the time steps assigned to each input. However, for simplicity, in this work, we assume that inputs must \textit{precede} one another.

We identify the family of problems where $n=1$ as \textit{Online Static GR (OSGR)} problems. They are static in the sense that in all instances, the set of goals remains the same. Later in this section, we will show that many GR solutions solve OSGR problems rather than their dynamic counterparts. Figure \ref{fig:GRAML} illustrates OSGR and ODGR problems. 
The next section reviews recent work, accompanied by an evaluative analysis 
associated with each methodology in light of the novel definitions.

\subsection{Model-Based GR (MBGR)}
In MBGR, the recognition process entails utilizing a pre-defined model that encompasses the characteristics of an environment along with the actions that can be executed within that environment~\citep{mirsky2021introduction, meneguzzi2021survey, masters2021goal}.
Traditional MBGR often exploits planning and parsing techniques \citep{avrahami2005fast,geib2009probabilistic,geib2009delaying, RamirezGeffner09, mirsky2016slim, son2016solving}. 
A major limitation of such approaches is their inflexibility in dynamic domains and that a complete model is crucial to solving the GR problem. Such algorithms require complex computation processes to calculate the likelihood of the goal given the observations. These computations are required for every given observation sequence, making these solutions less suitable for real-time performance. 

\citet{baker2009planning} and \citet{ramirez2011goal} propose approaches to perform GR over MDPs and Partially Observable MDPs (POMDPs) using planners, which generally incur lower computational costs than training RL agents. 
However, it is worth noting that MDP and POMDP planners may not scale as effectively as classical planners. The use of MDPs enables a framework to incorporate uncertainty modeling and, despite being model-based, opens a door for employing RL to solve GR. These approaches require significant computation for every observation sequence at Inference Time.

Several MBGR approaches focus on shortening the Inference Time by using pre-processing. For example, in metric-based goal recognition, distances between potential states are computed in the Domain Learning Time \citep{smith2015fast} or in the Goals Adaptation Time \citep{masters2017cost,vered2017heuristic,mirsky2019new}.

\subsection{Model-Free GR (MFGR)}
A Model-Free GR problem (MFGR)~\citep{geffner2018modelfree} is one in which the recognizer does not have access to the underlying model that describes the properties and dynamics of the environment. 
While some approaches learn the model dynamics and then employ MBGR methods~\citep{asai2018classical,amado2018goal}, other approaches perform GR directly, without learning the model of the world. Among these approaches, the main techniques include leveraging RL~\citep{amado2022goal} or employing deep neural networks to perform GR using a classification network~\citep{min2014deep,borrajo2020goal,chiari2023goal}. These methods 
typically have no Domain Learning Time, as the learning process depends on the set of goals, a fairly long Goals Adaptation Time, and a short Inference Time when given an observation sequence. 

 Different approaches have been proposed for learning an MFGR classification model using languages like PDDL~\citep{geib2018learning, maynard2019cost, borrajo2020goal, min2014deep,min2016player,fang2023real}. These approaches aim for generality by employing a simple domain file that specifies actions using labels and defines a state space based on a finite set of fluents (facts). \cite{chiari2023goal} proposed a framework (termed GRnet) that generalizes to any set of goals without additional learning, presenting a model-free solution suitable for environments with changing goals. 
 Compared to the other approaches, GRnet excels in ODGR, 
 demonstrating immediate adaptation to new goal sets and fast inference for new observations through a forward pass of the trained network. While this approach is promising, its reliance on fluent enumeration is critical for its success. 

A separate line of work uses MDPs instead of planning languages to model the environment. 
\citet{amado2022goal} presented an MDP-based method termed GR as RL, which trains RL agents during Domain Adaptation Time and computes the likelihood of such agents to produce the sequence of observations given at Inference Time. In this method, the sharing of information across various GR problems constructed by different observation traces with identical domain theories and goals is inherent. This sharing facilitates short 
 Inference Time. This approach integrates the domain theory with the problem instance by incorporating the policies of agents for each goal within their domain theory. The fact that learning typically occurs only after receiving the domain theory along with a set of goals limits the framework to recognizing goals only from this set. 
  This framework demonstrates high effectiveness in an OSGR setting, 
  However, as these approaches are not designed to handle dynamic environments if employed in an ODGR 
context, every new set of goals necessitates expensive re-training of agents for each distinct goal. 

%

\section{Dynamic Goal Recognition using Transfer Learning}

To tackle ODGR problems, we introduce a general algorithm based on transfer learning between source and target RL tasks~\citep{taylor2009transfer}. It facilitates learning in new tasks by transferring relevant parts from the policies of the source tasks. This algorithm selects specific goals, $G_b$, after receiving the domain theory $T$, and trains agents with policies $\Pi_{G_b}$ for each of them as part of the \hyperref[dlt]{Domain Learning time}. Upon receiving a set of goals $G_d$, the algorithm creates a set of policies $\Pi_{G_d}$ for each new goal by leveraging transfer learning from the base goals' trained agents, as part of the \hyperref[gat]{Goals Adaptation time}. When given an observation trace of an actor, the algorithm returns the most likely goal by using distance metrics, as part of the \hyperref[it]{Inference time}.
For convenience, we refer to the original set of goals, $G_b$, and the policies of their agents, $\Pi_{G_b}$, as \textit{base goals} and \textit{base policies}. Similarly, the new goals, $G_d$, in the adaptation phase are referred to as  \textit{dynamic goals}, and the policies generated for them, $\Pi_{G_d}$, are called \textit{dynamic policies}.

\begin{minipage}{0.46\textwidth}
\begin{algorithm}[H]
\caption{DomainLearningPhase}
\begin{algorithmic}[1]
\Require{$T = \langle S, A \rangle$ - a domain theory}
\State $G_b = \text{SelectBaseGoals}(T)$
\For{$g \in G_b$}
\State $q_g = \text{Learn}(T,g)$
\EndFor
\State $Q_{G_b} = \{q_g\}_{g\in G_b}$
\State $T_{G_b} = \langle S, A, Q_{G_b} \rangle$
\State \Return $T_{G_b}$
\end{algorithmic}
\label{alg:Domain_Learning_Phase}
\end{algorithm}

\end{minipage}
\hfill
\begin{minipage}{0.46\textwidth}
\begin{algorithm}[H]
\caption{GoalsAdaptationPhase}
\begin{algorithmic}[1]
\Require{$T_{G_b} = \langle S, A, Q_{G_b}=\{q_g\}_{g\in G_b} \rangle$}
\Require{$g_d \in G_d$: a set of new goals}
\For{$g \in G_d$}
\State $q_g = \text{Transfer}(T_{G_b} ,g)$
\EndFor
\State $Q_{G_d} = \{q_g\}_{g\in G_d}$
\State $T_{G_d} = \langle S, A, Q_{G_d} \rangle$
\State \Return $T_{G_d}$
\end{algorithmic}
\label{alg:Goals_Adaptation_Phase}
\end{algorithm}
\end{minipage}

\begin{algorithm}[ht]
\caption{InferencePhase}
\begin{algorithmic}[1]
\Require{$T_{G_c} = \langle S, A, Q_{G_c} \rangle, Q_{G_c} = \{q_g\}_{g\in G_c}$ - State and action spaces, and Q-functions per goal ($g_c$)}
\Require{$O$: an observation sequence (a sequence of states and actions tuples) $(\langle s_0, a_0 \rangle, \ldots, \langle s_t, a_t \rangle)$, where t is the length of $O$}
\State Sample/receive set of dynamic goals $G_d$
\State $m_g^* \gets \infty$ \Comment{Init shortest distance}
\For{all $g$ in $G_c$} \Comment{Compute distances from $O$}
\State $q_g^O$ is the Q-function of goal $g$ from $\{q_g\}_{g\in G_c}$ for $O$ states and actions
\State $m_g \gets \text{DISTANCE}(q_g^O, O)$ \Comment{Use distance measure}
\If{$m_g \leq m_g^*$}
\State $g^* \gets g$ and $m_g^* \gets m_g$
\EndIf
\EndFor
\State \Return $g^*$
\end{algorithmic}
\label{alg:Inference_Phase}
\end{algorithm}

Goal Adaptation using Transfer Learning (GATLing) consists of three parts that are essential to solving the ODGR problem. The \textit{Domain Learning Time} consists of building a domain theory using Algorithm \ref{alg:Domain_Learning_Phase}. Then, the system receives inputs sequentially. If the input is a set of new goals, it creates a new domain theory using algorithm \ref{alg:Goals_Adaptation_Phase}, which uses transfer learning from the base policies to train agents to the new goals, with the objective of minimizing the \textit{Goals Adaption Time} in this phase. The framework keeps the most updated domain theory for future recognition, and if the input is a trace of observations, it performs recognition using the most updated domain theory by calling Algorithm \ref{alg:Inference_Phase}. 
Algorithm \ref{alg:Inference_Phase} takes the dynamic goals' Q-functions, compares them to the observation, and chooses the most likely dynamic goal using the DISTANCE metric, operations that typically require a short \textit{Inference Time}. 


\subsection{Implementation in a Navigational Domain}

We implement GATLing using a simple navigational domain without obstacles to show feasibility. We use the Gym MiniGrid navigational domain~\citep{chevalier2023minigrid} as a case study, in which we conduct a comparative analysis of the different techniques' respective performances.

We implement Algorithm \ref{alg:Domain_Learning_Phase} using Q-Learning, as suggested in the GRAQL framework of~\citet{amado2022goal}. As part of Algorithm \ref{alg:Goals_Adaptation_Phase}, for policy transfer, we assign utilities to states and actions by applying weights to the expected utilities of actions from states in the Q-functions of the base goals. The weights assigned to these Q-functions 
are determined by the similarity of the path from the state to the base goal and the path to the dynamic goal, as perceived at each state individually. The notion of similarity can also be interpreted as distance in navigational domains. In the navigational domain, we used the distance metric as a similarity metric (where a lower distance means greater similarity). We employed two different methods to calculate this distance:

\noindent \textbf{Static} weights are assigned between each pair of base on dynamic goal, according to their Euclidean distance. The weight assignment is considered static as the weight for each action utility remains unchanged regardless of the states' spatial proximity to the dynamic goal. However, this approach introduces challenges, notably because it applies equal weights to all Q-values for every state in relation to a dynamic goal. As a result, in certain states, there may be a tendency to move in directions contrary to those leading to the dynamic goal, deviating from the optimal policy. This issue often arises from the proximity of the dynamic goal to another foundational goal in the opposite direction within these states.
    
\noindent \textbf{Dynamic} assigns variable weights to each state's utility. This weight assignment is based on the cosine similarity between the trajectory from the current state at which the Q-function was being calculated to the dynamic goal and the trajectory from this state to each base goal. 
%
The equation that describes the dynamic approach based on cosine similarity is denoted as follows:
\vspace{-2mm}
\begin{equation}
\small
\text{Cosine\_Similarity}(s, g_b, g_d) = \frac{\text{traj}(s, g_b) \cdot \text{traj}(s, g_d)}{\| \text{traj}(s, g_b) \| \cdot \| \text{traj}(s, g_d) \|}
\end{equation}
\vspace{-6mm}

where $s$ is the current state, $g_b$ is a base goal, $g_d$ is a dynamic goal, and $\text{traj}(s, g_b)$ represents a 2D vector originating in $s$ and ending at $g_b$.
%
The cosine similarity between two trajectories ranges from -1 (perfectly opposite directions) to 1 (perfectly aligned directions), with 0 indicating orthogonality.
%


To combine the policies of the base goals into a single policy for the dynamic goal, we employed three different aggregation techniques of the Q-values: 

\noindent\textbf{Weighted Average (normalize)} is a fair and accurate approach, but slow: $\frac{\sum_{i=1}^{n} w_i \cdot Q_i(s,a)}{\sum_{j=1}^{n} w_j}$

\noindent \textbf{Softmax} is an accurate approach, but slow, and highly variable:$\frac{\sum_{i=1}^{n} e^{w_i }\cdot Q_i(s, a)}{\sum_{j=1}^{n} e^{w_j}}$ 

\noindent \textbf{Max (Hard Weighted Average)} is a greedy approach, fast but less accurate: $\max_{i=1}^{n} Q_i(s, a)$


\subsection{Empirical Evaluation}
In this section, we present the results of our experiments, which evaluate GATLing in different scenarios. Our evaluation focuses on four key evaluation metrics: accuracy, precision, recall, and F-score.
%
%
As the output of GATLing is a distribution over a set of goals rather than a label, we explain the implementation of each metric in Appendix~\ref{app:metrics}.
  %
%

Evaluating policy transfer can be complex, as it varies greatly given the domain's dynamics and the properties of the source policies. Instead of running batches of runs on an arbitrary subspace of policies, we focus our experiments on two clear and controlled use cases:
the first presents a completely smooth environment (without sudden changes in Q-values) and the second breaks this smoothness property.
Both experiments employed Q-learning for learning, and during the inference stage, we leveraged KL-divergence to compare each policy with a pseudo-policy based on the observations reported by~\citet{amado2022goal}. The hyperparameters were set according to that work, and were consistent across both experiments: $\alpha = 1 \times 10^{-3}$, $\epsilon = 1 \times 10^{-8}$, $\gamma = 0.99$, and $Episodes = 10^7$. 
The base goals used for the first experiment were (1,6), (6,6), and (6,1), and for the second one (1,7), (7,7), and (7,1). The results are reported for partial traces of 0.1, 0.3, and 0.5, as traces with a completeness score of 0.7 or higher achieved a 100\% success rate across all four metrics in all our experiments, for both GRAQL and GATLing. (fuller traces showed absolute success for both GRAQL and GATLing). 
We also implemented a scaling factor that prioritizes and emphasizes each state's highest-ranked action (or actions) while reducing the Q-values of other actions in the same state. Appendix~\ref{app:scaling} shows an example of the Q-table generated with and without this scaling.

\begin{table} {
\begin{center}
\scriptsize
\centering
\begin{tabular}{cc|cccccccc|}
\cline{3-10}
\multicolumn{2}{c|}{\multirow{2}{*}{}} &
  \multicolumn{8}{c|}{Empty 8x8 Minigrid} \\ \cline{3-10} 
\multicolumn{2}{c|}{} &
  \multicolumn{4}{c|}{GATLing} &
  \multicolumn{4}{c|}{GRAQL} \\ \hline
\multicolumn{1}{|c|}{OBS} &
  Goals &
  \multicolumn{1}{c|}{Acc} &
  \multicolumn{1}{c|}{Prec} &
  \multicolumn{1}{c|}{Rec} &
  \multicolumn{1}{c|}{F-score} &
  \multicolumn{1}{c|}{Acc} &
  \multicolumn{1}{c|}{Prec} &
  \multicolumn{1}{c|}{Rec} &
  F-score \\ \hline
\multicolumn{1}{|c|}{\multirow{3}{*}{0.1}} &
  2 &
  \multicolumn{1}{c|}{1.0\textpm 0.0} &
  \multicolumn{1}{c|}{0.95\textpm 0.15} &
  \multicolumn{1}{c|}{0.95\textpm 0.15} &
  \multicolumn{1}{c|}{0.95\textpm 0.15} &
  \multicolumn{1}{c|}{1.0\textpm 0.0} &
  \multicolumn{1}{c|}{1.0\textpm 0.0} &
  \multicolumn{1}{c|}{1.0\textpm 0.0} &
  1.0\textpm 0.0 \\
\multicolumn{1}{|c|}{} &
  3 &
  \multicolumn{1}{c|}{0.95\textpm 0.15} &
  \multicolumn{1}{c|}{0.9\textpm0.2} &
  \multicolumn{1}{c|}{0.9\textpm 0.2} &
  \multicolumn{1}{c|}{0.9\textpm 0.2} &
  \multicolumn{1}{c|}{1.0\textpm 0.0} &
  \multicolumn{1}{c|}{1.0\textpm 0.0} &
  \multicolumn{1}{c|}{1.0\textpm 0.0} &
  1.0\textpm 0.0 \\
\multicolumn{1}{|c|}{} &
  4 &
  \multicolumn{1}{c|}{0.9\textpm 0.2} &
  \multicolumn{1}{c|}{0.8\textpm0.4} &
  \multicolumn{1}{c|}{0.8\textpm 0.4} &
  \multicolumn{1}{c|}{0.8\textpm 0.4} &
  \multicolumn{1}{c|}{0.95\textpm 0.15} &
  \multicolumn{1}{c|}{0.9\textpm 0.2} &
  \multicolumn{1}{c|}{0.9\textpm 0.2} &
  0.9 \textpm 0.2 \\ \hline
\multicolumn{1}{|c|}{\multirow{3}{*}{0.3}} &
  2 &
  \multicolumn{1}{c|}{1.0\textpm 0.0} &
  \multicolumn{1}{c|}{1.0\textpm 0.0} &
  \multicolumn{1}{c|}{1.0\textpm 0.0} &
  \multicolumn{1}{c|}{1.0\textpm 0.0} &
  \multicolumn{1}{c|}{1.0\textpm 0.0} &
  \multicolumn{1}{c|}{1.0\textpm 0.0} &
  \multicolumn{1}{c|}{1.0\textpm 0.0} &
  1.0\textpm 0.0 \\
\multicolumn{1}{|c|}{} &
  3 &
  \multicolumn{1}{c|}{1.0\textpm 0.0} &
  \multicolumn{1}{c|}{1.0\textpm 0.0} &
  \multicolumn{1}{c|}{1.0\textpm 0.0} &
  \multicolumn{1}{c|}{1.0\textpm 0.0} &
  \multicolumn{1}{c|}{1.0\textpm 0.0} &
  \multicolumn{1}{c|}{1.0\textpm 0.0} &
  \multicolumn{1}{c|}{1.0\textpm 0.0} &
  1.0\textpm 0.0 \\
\multicolumn{1}{|c|}{} &
  4 &
  \multicolumn{1}{c|}{0.95\textpm 0.15} &
  \multicolumn{1}{c|}{0.95\textpm 0.15} &
  \multicolumn{1}{c|}{0.95\textpm 0.15} &
  \multicolumn{1}{c|}{0.95\textpm 0.15} &
  \multicolumn{1}{c|}{1.0\textpm 0.0} &
  \multicolumn{1}{c|}{0.95\textpm 0.15} &
  \multicolumn{1}{c|}{0.95\textpm 0.15} &
  0.95 \textpm 0.15 \\ \hline
\multicolumn{1}{|c|}{\multirow{3}{*}{0.5}} &
  2 &
  \multicolumn{1}{c|}{1.0\textpm 0.0} &
  \multicolumn{1}{c|}{1.0\textpm 0.0} &
  \multicolumn{1}{c|}{1.0\textpm 0.0} &
  \multicolumn{1}{c|}{1.0\textpm 0.0} &
  \multicolumn{1}{c|}{1.0\textpm 0.0} &
  \multicolumn{1}{c|}{1.0\textpm 0.0} &
  \multicolumn{1}{c|}{1.0\textpm 0.0} &
  1.0\textpm 0.0 \\
\multicolumn{1}{|c|}{} &
  3 &
  \multicolumn{1}{c|}{1.0\textpm 0.0} &
  \multicolumn{1}{c|}{1.0\textpm 0.0} &
  \multicolumn{1}{c|}{1.0\textpm 0.0} &
  \multicolumn{1}{c|}{1.0\textpm 0.0} &
  \multicolumn{1}{c|}{1.0\textpm 0.0} &
  \multicolumn{1}{c|}{1.0\textpm 0.0} &
  \multicolumn{1}{c|}{1.0\textpm 0.0} &
  1.0\textpm 0.0 \\
\multicolumn{1}{|c|}{} &
  4 &
  \multicolumn{1}{c|}{1.0\textpm 0.0} &
  \multicolumn{1}{c|}{1.0\textpm 0.0} &
  \multicolumn{1}{c|}{1.0\textpm 0.0} &
  \multicolumn{1}{c|}{1.0\textpm 0.0} &
  \multicolumn{1}{c|}{1.0\textpm 0.0} &
  \multicolumn{1}{c|}{1.0\textpm 0.0} &
  \multicolumn{1}{c|}{1.0\textpm 0.0} &
  1.0\textpm 0.0 \\ \hline
\multicolumn{2}{|c|}{Average} &
  \multicolumn{1}{c|}{0.97 \textpm 0.05} &
  \multicolumn{1}{c|}{0.95 \textpm 0.1} &
  \multicolumn{1}{c|}{0.95 \textpm 0.1} &
  \multicolumn{1}{c|}{0.95 \textpm 0.1} &
  \multicolumn{1}{c|}{0.99 \textpm 0.01} &
  \multicolumn{1}{c|}{0.98 \textpm 0.03} &
  \multicolumn{1}{c|}{0.98 \textpm 0.03} &
  0.98 \textpm 0.03 \\ \hline

\end{tabular}%
\caption{
       A comparison of dynamic GATLing and GRAQL in Experiment 1. The setup varies by the number of goals and observability level (OBS). We present the accuracy (Acc), Precision (Prec), Recall (Rec), and F-score . Cells show the average result (± standard dev.) over 10 executions.
    }
    \label{tab:results_8x8}
\end{center}}
\end{table}

\begin{table*}[]{%
\scriptsize
\centering
\begin{tabular}{cc|cccccccc|}
\cline{3-10}
\multicolumn{2}{c|}{\multirow{2}{*}{}} &
  \multicolumn{8}{c|}{Simple Crossing (with 2 walls) 9x9 Minigrid} \\ \cline{3-10} 
\multicolumn{2}{c|}{} &
  \multicolumn{4}{c|}{GATLing} &
  \multicolumn{4}{c|}{GRAQL} \\ \hline
\multicolumn{1}{|c|}{OBS} &
  Goals &
  \multicolumn{1}{c|}{Acc} &
  \multicolumn{1}{c|}{Prec} &
  \multicolumn{1}{c|}{Rec} &
  \multicolumn{1}{c|}{F-score} &
  \multicolumn{1}{c|}{Acc} &
  \multicolumn{1}{c|}{Prec} &
  \multicolumn{1}{c|}{Rec} &
  F-score \\ \hline
\multicolumn{1}{|c|}{0.1} &
  2 &
  \multicolumn{1}{c|}{0.9\textpm 0.2} &
  \multicolumn{1}{c|}{0.9\textpm 0.2} &
  \multicolumn{1}{c|}{0.9\textpm 0.2} &
  \multicolumn{1}{c|}{0.9\textpm 0.2} &
  \multicolumn{1}{c|}{0.8\textpm 0.4} &
  \multicolumn{1}{c|}{0.8\textpm 0.4} &
  \multicolumn{1}{c|}{0.8\textpm 0.4} &
  0.8\textpm 0.4 \\ \hline
\multicolumn{1}{|c|}{0.3} &
  2 &
  \multicolumn{1}{c|}{1.0\textpm 0.0} &
  \multicolumn{1}{c|}{1.0\textpm 0.0} &
  \multicolumn{1}{c|}{1.0\textpm 0.0} &
  \multicolumn{1}{c|}{1.0\textpm 0.0} &
  \multicolumn{1}{c|}{1.0\textpm 0.0} &
  \multicolumn{1}{c|}{1.0\textpm 0.0} &
  \multicolumn{1}{c|}{1.0\textpm 0.0} &
  1.0\textpm 0.0 \\ \hline
\multicolumn{1}{|c|}{0.5} &
  2 &
  \multicolumn{1}{c|}{1.0\textpm 0.0} &
  \multicolumn{1}{c|}{1.0\textpm 0.0} &
  \multicolumn{1}{c|}{1.0\textpm 0.0} &
  \multicolumn{1}{c|}{1.0\textpm 0.0} &
  \multicolumn{1}{c|}{1.0\textpm 0.0} &
  \multicolumn{1}{c|}{1.0\textpm 0.0} &
  \multicolumn{1}{c|}{1.0\textpm 0.0} &
  1.0\textpm 0.0 \\ \hline
\multicolumn{2}{|c|}{Average} &
  \multicolumn{1}{c|}{0.96 \textpm 0.06} &
  \multicolumn{1}{c|}{0.96 \textpm 0.06} &
  \multicolumn{1}{c|}{0.96 \textpm 0.06} &
  \multicolumn{1}{c|}{0.96 \textpm 0.06} &
  \multicolumn{1}{c|}{0.93 \textpm 0.13} &
  \multicolumn{1}{c|}{0.93 \textpm 0.13} &
  \multicolumn{1}{c|}{0.93 \textpm 0.13} &
  0.93 \textpm 0.13 \\ \hline
\end{tabular}%
}
\caption{
        A comparison of dynamic GATLing and GRAQL in Experiment 2.
    }
    \label{tab:results_9x9}
\end{table*}

\subsubsection{Results}



Tables \ref{tab:results_8x8} and \ref{tab:results_9x9} show the results from Experiment 1 and Experiment 2, respectively. 
The dynamic approach significantly outperformed the \textit{static} approach, highlighting the importance of adaptability in ODGR. When using the dynamic approach, the softmax and normalized techniques were always able to effectively combine policies from base goals' agents into coherent policies for dynamic goals. In contrast, the max aggregation could not rank the true goal as the most likely one. We, therefore, show the performance results of the \textit{softmax} aggregation method with a \textit{dynamic} distance metric that performed best, leaving out the results using other approaches presented in this paper.
%
%
Regarding runtime, GATLing policy construction comes only after a Domain Learning Phase, in which runtime compares with GRAQL's. Similarly, the inference times are swift for both GATLing and GRAQL because both don't require any form of learning or planning. However, constructing a dynamic goal policy using the GATLing framework required only 1.56 seconds on average across 10 runs on a GPU. In contrast, learning a single goal with the GRAQL approach took significantly longer, averaging 485.45 seconds across 10 runs on the same GPU, emphasizing the potential of transfer learning in requiring shorter goal adaptation times. While this evaluation cannot precisely determine the performance of different approaches due to implementation variability, it provides some insight into the potential benefits of considering ODGR problems. 

\section{Conclusion}
This paper presented the novel problem of ODGR, which emphasizes the time in which different parts of the input are received, and the potential gain from policy transer. It formally defines ODGR problems and how they compare to most existing GR problems. Then, the paper provides an overview of existing GR algorithms and their computational effort distribution according to ODGR's components. Lastly, the paper introduces GATLing, a general algorithm for performing ODGR in 2D navigational domains, with an implementation using two distance metrics and three aggregation techniques for the policy ensemble. 
Our experiments demonstrate that in navigational domains without obstacles, GATLing offers effective ODGR capabilities, with accurate recognition performance and significantly reduced run-time compared to existing RL recognition approaches. 
%

While GATLing provides a foundation for future algorithms, there are several clear areas for potential improvements. First, the proposed aggregation techniques are expected to fail in domains that consist of rapid changes in Q values. Moreover, it is not straightforward how this approach can be extended to continuous state and action spaces. Nevertheless, by avoiding the lengthy policy learning phases. 
GATLing ensures that the recognition process remains efficient and scalable, even in environments where goals may emerge or evolve. 
%
As the field of data-driven GR develops, we anticipate that the need for faster goal learning will increase. This research is expected to contribute to developing adaptable and efficient GR systems in dynamic, changing environments.

\bibliography{rlc}
\bibliographystyle{rlc}

\newpage
\appendix
\section{Evaluation Metrics Definition}
\label{app:metrics}
As the output of our GATLing is a distribution over a set of goals rather than a label, we explain how we implemented each measure:
\begin{itemize}
    \item \textbf{Accuracy} measures the proportion of correctly recognized dynamic goals. A goal is considered correctly recognized if it has the highest likelihood of being the actor's goal in comparison to other goals.
    \item   \textbf{Precision} indicates how many of the most likely goals were the actor's goal.
    \item \textbf{Recall} indicated how many of the actor's true goals were correctly identified. 
    \item  \textbf{F-score} is the harmonic mean of Precision and Recall.
\end{itemize}

\section{Scaling Factor for Q-function Aggregation}
\label{app:scaling}
\begin{figure}[ht]
\begin{center}
  \includegraphics[width=0.23\textwidth]{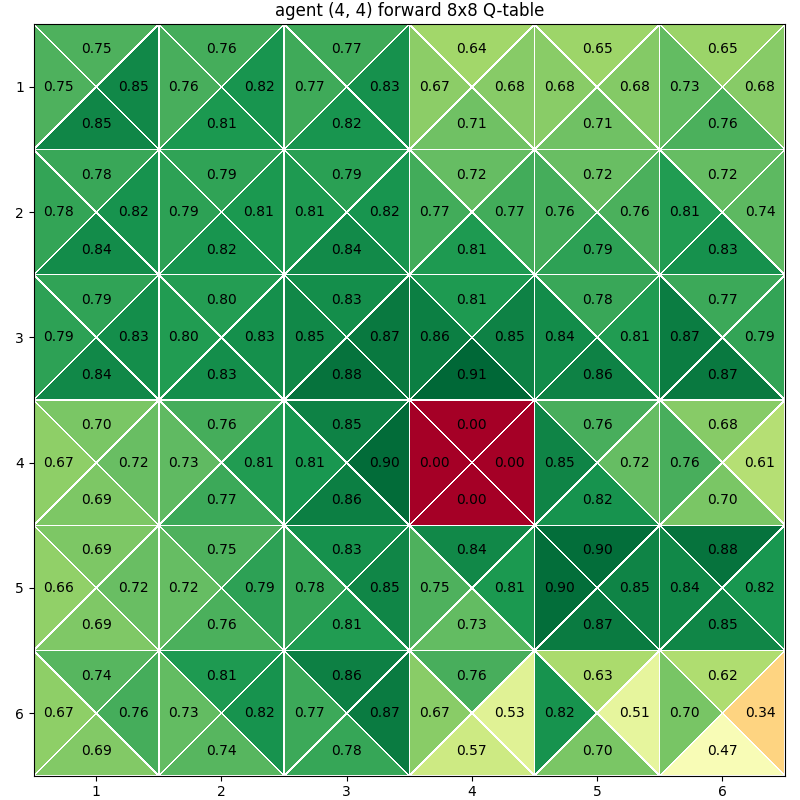}
  \includegraphics[width=0.23\textwidth]{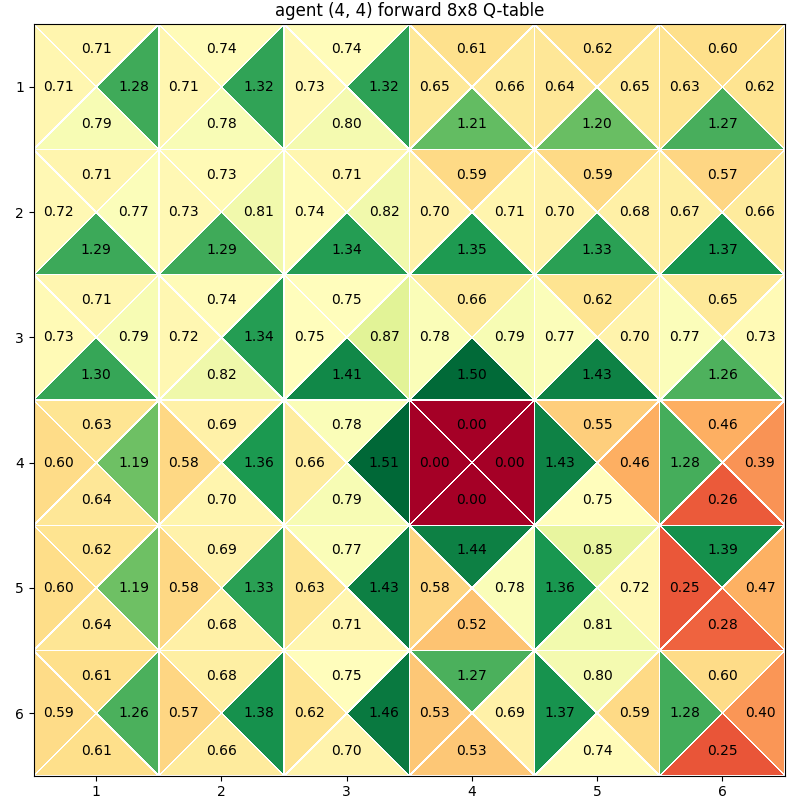}
  \caption{The heuristic Q-table 
  generated from Q-tables whose base goals are at (1,6), (6,1), and (6,6) using cosine similarity and softmax (without scaling) at the $8 \times 8$ environment. Each cell shows the Q-values for taking the actions up, right, down, and left. The left (right) grid shows the outcome Q-table without (with) scaling.}
  \label{fake_8x8_without_scaling}
\end{center}
\end{figure}

Sometimes, the desired action may not have the highest Q-value after aggregation. This phenomenon can sometimes be attributed to how two base goals' policies bias the dynamic policy in a slightly wrong direction. To mitigate this issue, we implemented a scaling factor that prioritizes and emphasizes each state's highest-ranked action (or actions) while reducing the Q-values of other actions in the same state. Figures \ref{fake_8x8_without_scaling} (left) and \ref{fake_8x8_without_scaling} (right) show the crafting of a dynamic goal (4,4) Q-table for a $8 \times 8$ Minigrid environment without obstacles. The experimental setup is the same as experiment 1. These scaling methods fine-tune the Q-values for particular actions for every state to ensure consistent and optimal behavior of the agent.

\end{document}